\documentclass[conference]{ieeeconf}

\IEEEoverridecommandlockouts                              % This command is only needed if 
\usepackage{graphicx} % for pdf, bitmapped graphics files
\usepackage{amsmath} % assumes amsmath package installed
\usepackage[
backend=biber,
style=numeric,
citestyle=ieee
]{biblatex}
\usepackage[ruled,vlined]{algorithm2e}
\usepackage{array}
\usepackage{tabularx}
\title{\LARGE \bf
Track based Offline Policy Learning for Overtaking Maneuvers with Autonomous Racecars}

\author{Jayanth Bhargav$^{1}$, Johannes Betz$^{1}$, Hongrui Zheng $^{1}$, Rahul Mangharam$^{1}$
\thanks{$^{1}$University of Pennsylvania, Department of Electrical and Systems Engineering,
School of Engineering and Applied Sciences
        {\tt\small joebetz@seas.upenn.edu}}
}

\begin{document}

\maketitle
\thispagestyle{empty}
\pagestyle{empty}

%%%%%%%%%%%%%%%%%%%%%%%%%%%%%%%%%%%%%%%%%%%%%%%%%%%%%%%%%%%%%%%%
%%%%%%%%%%%%%%%%%%%%%%%%%%%%%%%%%%%%%%%%%%%%%%%%%%%%%%%%%%%%%%%%
%%%%%%%%%%%%%%%%%%%%%%%%%%%%%%%%%%%%%%%%%%%%%%%%%%%%%%%%%%%%%%%%

\begin{abstract}
The rising popularity of driver-less cars has led to the research and development in the field of autonomous racing, and overtaking in autonomous racing is a challenging task. Vehicles have to detect and operate at the limits of dynamic handling and decisions in the car have to be made at high speeds and high acceleration. One of the most crucial parts in autonomous racing is path planning and decision making for an overtaking maneuver with a dynamic opponent vehicle. In this paper we present the evaluation of a track based offline policy learning approach for autonomous racing. We define specific track portions and conduct offline experiments to evaluate the probability of an overtaking maneuver based on speed and position of the ego vehicle. Based on these experiments we can define overtaking probability distributions for each of the track portions. Further, we propose a switching MPCC controller setup for incorporating the learnt policies to achieve a higher rate of overtaking maneuvers. By exhaustive simulations, we show that our proposed algorithm is able to increase the number of overtakes at different track portions.

\end{abstract}

%%%%%%%%%%%%%%%%%%%%%%%%%%%%%%%%%%%%%%%%%%%%%%%%%%%%%%%%%%%%%%%%
%%%%%%%%%%%%%%%%%%%%%%%%%%%%%%%%%%%%%%%%%%%%%%%%%%%%%%%%%%%%%%%%
%%%%%%%%%%%%%%%%%%%%%%%%%%%%%%%%%%%%%%%%%%%%%%%%%%%%%%%%%%%%%%%%
\section{Introduction}
\label{sec:introduction}
\subsection{Autonomous Racing}

Autonomous racing has become popular over the recent years and competitions like Roborace \cite{Betz2018}  or the Indy Autonomous Challenge as well as with small-scale racecars like F1Tenth \cite{okelly2020f1tenth}  provide platforms for evaluating autonomous driving software. The overall goal of all these competitions is that researchers and engineers can develop algorithms that operate at the vehicles edge: High speeds, high accelerations, high computation power, adversarial environments. Similar to normal racing series like Formula 1 the development of algorithms for autonomous racing generate trust in the field of autonomous driving and enables the development of advanced autonomous driving algorithms.\\
The algorithms that were developed in the field of autonomous racing so far are mostly focusing on single vehicle only that try to achieve a human-like lap time. The field of high dynamic overtaking maneuver with dynamic opponents was less displayed so far. In addition, achieving a human-like behavior (e.g. like a Formula 1 race driver) that makes the decision about an overtaking maneuver and executes a secure and reliable maneuver at high speeds is still an unsolved question. 

\subsection{Contributions}
\label{subssec:contributions}

Based on the state of the art, in this paper we present an offline policy learning for overtaking maneuvers in autonomous racing. This work has two primary contributions:
\begin{enumerate}
    \item We provide a design of experiment (DoE) for an offline driven policy learning approach by track discretization. Based on the specific track we provide a discretization into turns and vary the position and velocity of the ego vehicle for these turns. With this DoE we create policies for the racecar that will enable a better overtaking maneuver. These policies learnt for different track portions are then integrated into path planning and control. We show that this approach is effective in increasing the number of overtakes at different track portions. 
    \item The overtaking policies learnt from the experiments are then incoparated into a MPCC planner. We propose a switched MPCC controller which combines a receding horizon control algorithm and specific driving behaviours like driving on the left side and driving on the right side. These modes restrict the action space for agent/vehicle and helps to enforce practical overtaking strategies like overtake from inside of track or stay on outside of the track at particular turns.
\end{enumerate}
In summary, with the setup defined in this paper we can create more realistic and better overtaking maneuvers for autonomous race vehicles. It also provides the intuition and domain knowledge for tuning learning-based planners and controllers.

%%%%%%%%%%%%%%%%%%%%%%%%%%%%%%%%%%%%%%%%%%%%%%%%%%%%%%%%%%%%%%%%

\section{Design of Experiments}
\label{sec:experiments}
The main idea behind making better overtaking is to learn from the track. Each track has a specific layout and a combination of turns and straights. Each turn has a specific curvature and therefore allows the car to drive only a a maximum velocity. In addition, to drive with the maximum speeds the car needs to follow a specific trajectory that leads through the turns.\\
We propose now an offline experiment setup which will create a detailed insights on specific track portions and the possibility to overtake in specific turns. With these offline experiments it is possible to create track-based policies that can be used to  enhance the MPCC planner by integrating these policies into its cost-function and constraints. 

\subsection{Track Portions}
In the first step we are defining the track portions that we will examine. We will use here four track portions that the most common kinds of curves/turns found on racetracks:
\begin{itemize}
    \item \textit{Straight}: This is a straight part of the track  which has  curvature. It is meant to let drivers push the cars to their speed limits. The straight is the quickest and easiest track portion for overtaking maneuvers.
    
    \item \textit{Sweeper}: The sweeper curve is a gradual, long and large turn. These curves are usually very wide and have high banking which makes it easier for race cars to pass by. The cars reach high velocities here and the overtaking can be done on the inside or the outside.
    
    \item \textit{Hairpin}: The hairpin has the biggest curvature and is mostly the tightest turn on a track. Here we will see the car slow down to and make very sharp turns. overtaking maneuvers happen here mostly before entering the hairpin.
    
    \item \textit{Chicane}: The chicane is a straight modified into a slight S shape. The velocities that the car is reaching here are varying from high to low based on the racetrack. Overtaking maneuvers happen here less often because its less space due to the steering maneuvers.
\end{itemize}

\begin{figure}[h!]
    \includegraphics[scale=0.22]{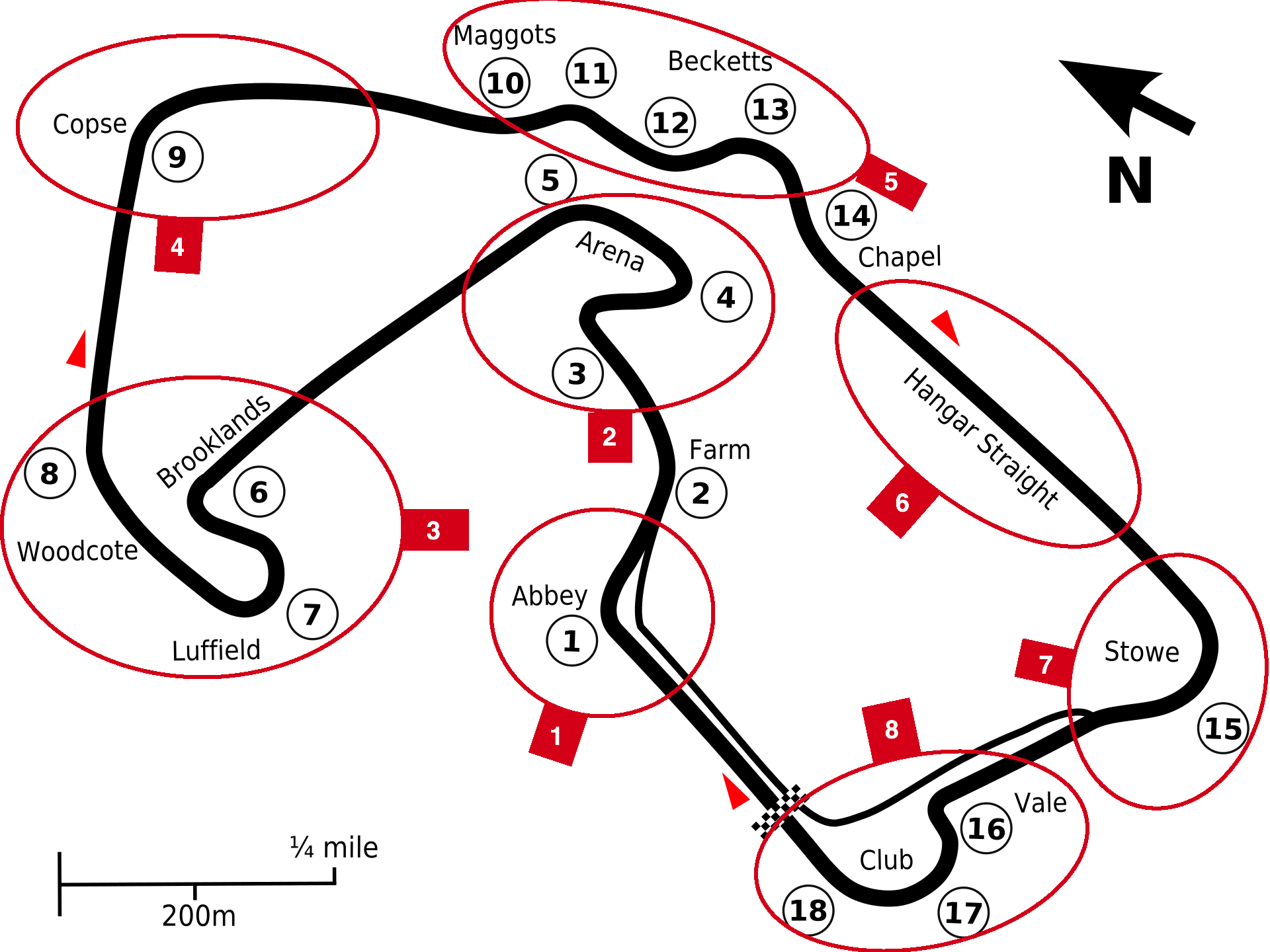}
    \caption{Silverstone Circuit}
    \label{figure:silver}
\end{figure}

As a showcase in this paper we are using the Silverstone racetrack. The choice of Silverstone track for our experiments is backed by the fact that this track layout has all of the above types of curves which is challenging for an overtaking maneuver. In figure \ref{figure:silver} we display 9 different track portions on the Silverstone track that are marked with labels 1 to 8. The track portions are defined by $\mathbf{T} = \{\tau | \tau \in (1,2,3,4,5,6,7,8)\}$ and are  categorized in the four track segment types: 
\begin{itemize}
    \item \textit{Straight}: 6
    \item \textit{Sweeper}: 1, 4, 7
    \item \textit{Hairpin}: 2, 3, 8
    \item \textit{Chicane}: 5
\end{itemize}

\begin{figure}[h!]
    \includegraphics[scale=0.25]{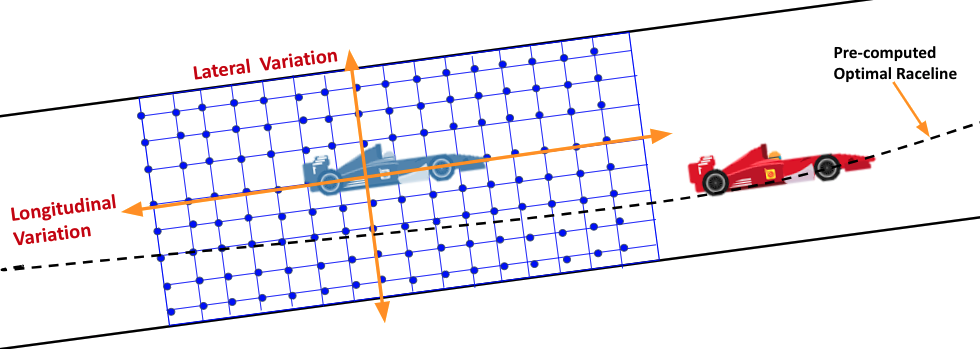}
    \caption{Ego vehicle (blue car) starting behind the opponent (red car) vehicle. The opponent vehicle follows a pre-computed raceline, the position and starting velocity of the ego vehicle are varied.}
    \label{track_variations}
\end{figure}

subsection{Sampling based Trajectory Rollouts}
To examine these defined track portions we setup an offline simulation that varies different parameters visualized in figure \ref{track_variations}. 

The simulation setup consists of two agents: One ego and one opponent vehicle. The opponent vehicle follows a pre-computed race line based on \cite{Heilmeier2019} and is non-interactive. The ego vehicle is starting behind the opponent vehicle and is an intelligent agent that planning its path based on the MPCC algorithm.
For every track portion $\tau \in \mathbf{ T }$, a uniformly sampled set of positions $\mathcal{P}: \mathcal{X} X \mathcal{Y} \subset \mathbf{R}^2$ are chosen as the starting position for the ego. The obstacle vehicle however always tracks the pre-computed race line. The obstacle vehicle speed is varied as $v_{obs} = v_{baseline} * (1+s)$ where $s \in \{-0.2, 0, +0.2\}$, $v_{baseline}$ being the speed of the obstacle from the pre-computed curvature optimal race line.

The agents are initialised with these positions and set to start the simulation. The ego vehicle synthesises dynamic trajectories based on the MPCC planner with obstacle avoidance. A fully observable model is used for the ego vehicle i.e. the ego vehicle will have the information of the track portion $\tau$ which it is driving in and the current state of the obstacle $X_{obs} = [x_{obs},y_{obs},\phi_{obs}]$

In this setup, we conduct exhaustive simulations based on the following parameter variations: 
\begin{itemize}
    \item Lateral offset: The position of the ego vehicle is varied lateral across the track with an offset from the centerline.
    \item Longitudinal Offset: The position of the ego vehicle is varied longitudinal along the centerline of the track.
    \item Opponent Speed Change: The opponent speed is varied with $\pm 20 \%$ from baseline
\end{itemize}

With this kind of simulations we have now the possibilty to examine if a specific position in one of the defined track portions has an advantage for an overtaking maneuver or not. 

It is highly expected that the ego vehicle will succeed in an overtaking maneuver when the obstacle speed is $20\%$ lower than its baseline. This establishes the fact that speed advantage always helps in overtaking (e.g. DRS zones in F1). The next set of parameters that influence the overtaking maneuver is the position. In a convoluted track like the Silverstone circuit we can display if starting off at a specific position enables a higher chance of an overtaking maneuver. For each track portion, we define four regions of interest: $\mathcal{R}_1$, $\mathcal{R}_2$, $\mathcal{R}_3$ and $\mathcal{R}_4$. Starting positions of the ego vehicle are uniformly sampled in all the four regions to generate experimental data. 

%%%%%%%%%%%%%%%%%%%%%%%%%%%%%%%%%%%%%%%%%%%%%%%%%%%%%%%%%%%%%%%%
\section{Related Work}
\label{sec:related_work}

Although the state of the art displays tons of algorithms for path and behavioral planning of autonomous vehicles, the explicit algorithm development for autonomous race cars is just done in a small community. As part of the Roborace competition \cite{Betz2019} and \cite{Caporale2018} presented a planning an control system for real life autonomous racing cars. Both approaches focused on a holistic software architecture that is capable of dynamic overtaking. Nevertheless none of them realized a head to head race with the vehicles. As a part of the same competition, \cite{Buyval2017} presented a nonlinear model predictive control (NMPC) for racing. The overtaking strategy was implemented as a term in the objective function. The NMPC has the freedom to choose the side for an overtake and was mainly relying on the obstacles velocity to perform the overtaking maneuver. \\
Beside these classical control driven approaches, new machine learning algorithms were used to learn how to race. \cite{Weiss2020}  displayed an end-to-end framework for a computer game  (Formula One (F1) Codemasters) that is using Convolution Neural Network (CNN) integrated with Long Short-Term Memory (LSTM) to learn how to drive fast an reliable around a racetrack. In \cite{Schwarting2020} a reinforcement learning algorithm was presented that learns competitive visual control policies through self-play in imagination. Although this setup is not focusing on real vehicle dynamic behavior, it provided interesting strategies for multi vehicle interaction. \\
When it comes to multi vehicle racing, only a few researchers have considered the whole pipeline of path and behavioral planning in combination with the interactions the vehicle can make. \cite{Liniger2020} presented a non-cooperative game theory approach where autonomous racing is formulated as racing decisions as a non-cooperative nonzero-sum game. Liniger et al. displayed that different games can be modelled that achieve successfully different racing behaviors and generate interesting racing situations e.g. blocking and overtaking. Notomista et al. \cite{Notomista2020} considered a two-player racing game where the ego vehicle is based on a Sensitivity-ENhanced NAsh equilibrium seeking (SENNA) method, which uses an iterated best response algorithm in order to optimize for a trajectory in a two-car racing game. Focusing on collision avoidance, the SENNA methods exploits interactions between the ego and the opponent vehicle. Unfortunately this approach was only displayed in an unknown simulation environment and therefore its performance is unknown. Wang et. al  \cite{wang_game-theoretic_2021} proposed a nonlinear receding horizon game-theoretic planner for autonomous cars in competitive scenarios with other cars. By implementing the trajectory as a piecewise polynomial and incorporating bicycle kinematics into the trajectory the authors were able to show that the ego vehicle exhibits rich game strategies such as blocking, faking, and opportunistic overtaking maneuvers.

The state of the art displays that the autonomous racing community is focusing on integrating effective learning techniques and strategies into dynamic path and behavioral planning. Additionally, authors have displayed individual algorithms and methods that are trying to make the car faster, more reliable and more interactive \cite{Rosolia2020} \cite{Kabzan2019} \cite{Kapania2020}.
Majority of the literature is centered around extensions to existing path and behavioral planning approaches. Improvements in planning/control for overtaking maneuvers have not yet been explored rigorously. Techniques to optimize path planning and control algorithms to ensure a better overtaking maneuver have to addressed. 

%%%%%%%%%%%%%%%%%%%%%%%%%%%%%%%%%%%%%%%%%%%%%%%%%%%%%%%%%%%%%%%%
%%%%%%%%%%%%%%%%%%%%%%%%%%%%%%%%%%%%%%%%%%%%%%%%%%%%%%%%%%%%%%%%
%%%%%%%%%%%%%%%%%%%%%%%%%%%%%%%%%%%%%%%%%%%%%%%%%%%%%%%%%%%%%%%%

\section{Planning and Control Setup}
\label{sec:optimalcontrol}
The continuous time system dynamics described in Section \ref{sec:vehiclemodel} is used to develop a constrained optimal controller to steer the vehicle in the track. The optimal planner plans the path for a horizon of $N$ steps ahead, steers the vehicle with the first step, and again repeats the process for the specified amount of time. This is a modified form of the Model Predictive Controller (MPC).  \newline
\subsection{Model Predictive Contouring Control}
The Model Predictive Contouring Control (MPCC) problem defined in \cite{liniger2015optimization} is re-formulated into a finite-continuous time optimal control problem as follows: \newline
\vspace{-0.5cm}
\begin{center}
    
\begin{equation*}
\min \int_{0}^{T}\begin{bmatrix}
 \epsilon_{c}^{l i n}(t) &  \epsilon_{l}^{l i n}(t)
\end{bmatrix}\begin{bmatrix}
Q c & 0 & \\ 
0 & Q l & 
\end{bmatrix}\begin{bmatrix}
\epsilon_{c}^{l i n}(t) \\ 
\epsilon_{l}^{l i n}(t) 
\end{bmatrix} 
\end{equation*}

\vspace{-0.5cm}

\begin{equation*}
 - Q_{\theta} \dot{\theta}(t)+u^{T}(t) R u(t) d t
\end{equation*}

\vspace{-0.5cm}

\begin{equation*}
\text{s.t.} \quad   \begin{aligned}[t]
    \quad  \dot{x}=f(x, u, \Phi) \\
    \quad b_{l o w e r} \preceq x(t) \preceq b_{\text {upper }} \\
    \quad l_{\text {lower }} \preceq u(t) \preceq l_{\text {upper }} \\
    \quad  h(x, \Phi) \leq 0 
  \end{aligned}
\end{equation*}
\end{center}

given the system dynamics $f$ and the arclength parametrization of the contour (the track) $\Phi$. A single-track bicycle model is used. Section \ref{sec:vehiclemodel} contains the details of the vehicle dynamics. Here $x(t)$ denotes the system state, $u(t)$ the inputs to the system, $b$ the box constraints on the state, $l$ the box constraints on the input and $h$ captures the track boundary constraints. The state of the system is augmented with the advancing parameter $\theta$

$$
x=\left[\begin{array}{c}
x_{\text {model }} \\
\theta
\end{array}\right]
$$
and the virtual input $\dot{\theta}$ is appended to the inputs from the original system dynamics.
$$
u=\left[\begin{array}{c}
u_{\text {model }} \\
\dot{\theta}
\end{array}\right]
$$
The track boundary constraint is realized as a convex disk constraint.
$$
h(x, \Phi)=\left(x-x_{t}^{l i n}(\theta)\right)^{2}+\left(y-y_{t}^{l i n}(\theta)\right)^{2}-r_{\Phi}(\hat{\theta})^{2}
$$
Here $r_{\Phi}(\hat{\theta})$ is the half-width of the track at the last predicted arc length.

\begin{figure}[h!]
\centering
    \includegraphics[scale=0.32]{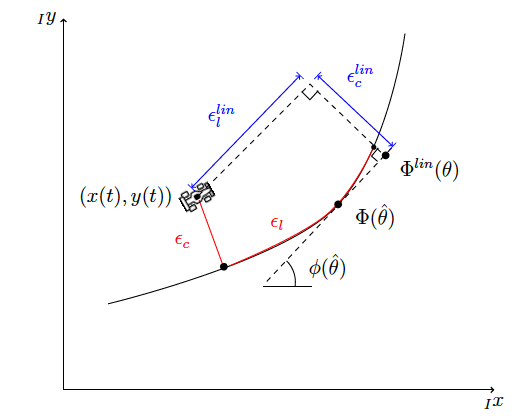}
    \caption{Linearized contouring and lag errors}
    \label{fig:track_lin}
\end{figure}

The linearized contouring error $\epsilon_{c}^{l i n}$ and lag error $\epsilon_{l}^{l i n}$ are computed as shown in Figure \ref{fig:track_lin}. To make the problem real-time feasible they are approximated by linearizing both them and the track around the previous solution $\theta$ as:
$$
\begin{aligned}
\Phi(\theta) &=\left[\begin{array}{l}
x_{t}(\theta) \\
y_{t}(\theta)
\end{array}\right] \approx \Phi(\hat{\theta})+\partial_{\theta} \Phi(\hat{\theta})(\theta-\hat{\theta}) \\
\end{aligned}
$$
$$
\begin{aligned}
\Rightarrow \Phi^{l i n}(\theta) &=\left[\begin{array}{l}
x_{t}(\hat{\theta})+\cos (\phi(\hat{\theta}))(\theta-\hat{\theta}) \\
y_{t}(\hat{\theta})+\sin (\phi(\hat{\theta}))(\theta-\hat{\theta})
\end{array}\right]
\end{aligned}
$$
this allows us to compute the errors
$$
\begin{aligned}
x_{t}^{l i n}(\theta) &=x(t)+\epsilon_{l}^{l i n} \cos (\phi(\hat{\theta}))+\epsilon_{c}^{l i n} \sin (\phi(\hat{\theta})) \\
y_{t}^{l i n}(\theta) 
\end{aligned}
$$
$$
\begin{aligned}
&=y(t)+\epsilon_{l}^{l i n} \sin (\phi(\hat{\theta}))-\epsilon_{c}^{l i n} \cos (\phi(\hat{\theta})) \\
\Leftrightarrow \\
\end{aligned}
$$
$$
\left\{\begin{matrix}
\epsilon_{l}^{l i n} =\cos (\phi(\hat{\theta}))\left(x_{t}^{l i n}(\theta)-x(t)\right)+\sin (\phi(\hat{\theta}))\left(y_{t}^{l i n}(\theta)-y(t)\right) \\
\epsilon_{c}^{l i n} =\sin (\phi(\hat{\theta}))\left(x_{t}^{l i n}(\theta)-x(t)\right)-\cos (\phi(\hat{\theta}))\left(y_{t}^{l i n}(\theta)-y(t)\right)\\ 

\end{matrix}\right.
$$
These approximations hold good if $\epsilon_{l} \approx 0$ and $\hat{\theta}-\theta \approx 0 .$ In practice the first can be incentivized by increasing $Q_{l}$ and the second by warmstarting the problem correctly.

The MPCC is optimizing to move the position of a virtual point $\theta(t)$ along the track to achieve as much progress as possible while steering the model of the vehicle to keep contouring and lag errors as small as possible. \newline

\subsection{Switched Model Predictive Contouring Control}
\begin{figure}[h!]
\begin{center}
    
    \includegraphics[scale=1.0]{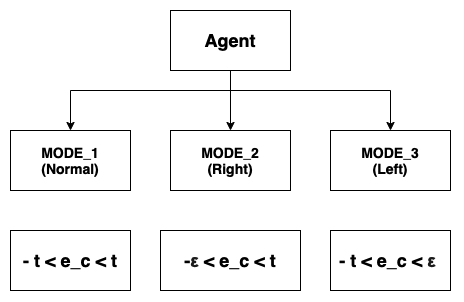}
    \caption{Switched MPCC Controller}
    \label{figure:switchedmpc}
    \end{center}
\end{figure}

The MPCC controller defined above ensures that the vehicle stays inside the track, drives fast and avoids obstacles. The track and obstacle constraints are realised as non-linear convex disk constraints. However, such a setup does not give enough control over achieving different driving behaviours like driving on the inside of a chicane (left side of the track), driving on the outside of a sweeper (right side of the track). To achieve more control over the path planning of the ego vehicle, a switched MPCC setup is proposed and displayed in figure \ref{figure:switchedmpc}. 

In this, the agent switches between different modes defined by different solver formulations. The overtaking possibilities for the MPCC in this paper are:

\begin{itemize}
    \item \textbf{Normal Mode: } In this, the search space for the N-horizon MPCC is the complete racetrack
    \item \textbf{Drive Right: } In this, the search space of the N-horizon MPCC is restricted only right half of the track. 
      \item \textbf{Drive Left: } In this, the search space of the N-horizon MPCC is restricted only left half of the track. 
\end{itemize}
The "Drive Right" and "Drive Left" modes are achieved by imposing additional constraints on the contouring error of the vehicle. The constraints of the optimisation problem are tweaked with slack variables (like $\epsilon$) to ensure that the planner does not get stuck into an in-feasibility loop leading to a crash. \newline
Note that many more of these modes can be defined and added to the MPCC. These modes can then integrate practical driving behaviours like ADAS mode (making tighter bounds on the problem and not letting the vehicle drive at its limits) and Push-to-Pass (giving a speed boost for a short window of time). \\
The MPCC control problem with a horizon of $N = 35$ is modeled as a quadratic problem (QP) with linear and non linear inequalities solved by efficient interior point solvers in FORCES \cite{Zanelli2017}. At each step of the simulation, the solver generates the optimal control input $u^*$ for a horizon of $N$. The first value $u^*(0)$ is used to steer the vehicle and the procedure is repeated. 

\subsection{Track Parametrization}

The center-line of the track is given in way-points (X-and Y-position). To implement MPCC an arc-length parametrization $\Phi$ is required. This is realized by interpolating the way-points using cubic splines with a cyclic boundary condition, and creating a dense lookup table with the track location and the linearization parameters. Note that in the optimization it is not practical to pass the full parametrization since it contains functions that the used solver has difficulties dealing with such as floor and modulo. Instead the linearization parameters are precomputed offline and passed at every stage.

%%%%%%%%%%%%%%%%%%%%%%%%%%%%%%%%%%%%%%%%%%%%%%%%%%%%%%%%%%%%%%%%%%%%%%%%%%%%%%%%
%%%%%%%%%%%%%%%%%%%%%%%%%%%%%%%%%%%%%%%%%%%%%%%%%%%%%%%%%%%%%%%%%%%%%%%%%%%%%%%%
%%%%%%%%%%%%%%%%%%%%%%%%%%%%%%%%%%%%%%%%%%%%%%%%%%%%%%%%%%%%%%%%%%%%%%%%%%%%%%%%

\section{Offline Policy Learning for Overtaking}
\label{sec:method}
Algorithm \ref{alg:policyalg} elucidates the offline experiment based policy learning developed in this paper.  \newline
$\mathcal{X}$, $\mathcal{Y}$ are the set of x and y coordinate offsets (expressed as percentage of track width) and $\mathcal{S} = \{-0.2, 0, +0.2\}$ is the speed offset expressed as a percentage change from the baseline obstacle speed.  

The obstacle update model is $g$, which is a pre-computed curvature optimal race line of the Silverstone track. 
One run of the algorithm populates the policy map $\Pi$ with the track regions for each of the 8 curves, having highest probability of overtakes. In total we are running a number of 576 experiments based on the 16 lateral, 12 longitudinal and 3 velocity variations for each of the 8 track portions.

\begin{algorithm}
    \label{alg:policyalg}
    \SetAlgoLined
    
    \SetKwFunction{FMain}{ MPCC Planner}
    \SetKwProg{Fn}{Function}{:}{}
    \Fn{\FMain{$X_{obs}$}}{
        Solve MPCC Problem defined in Section \ref{sec:optimalcontrol}
        \KwRet $u^*$\;
    }
      
    \SetKwFunction{FMain}{ Check Overtake}
    \SetKwProg{Fn}{Function}{:}{}
    \Fn{\FMain{$X_{obs}$, $X_{ego}$}}{
        Project $X_{obs}$, $X_{ego}$ as $s_1, s_2$ on track \\
         \uIf{$s_1 > s_2$}{
        overtake\_status = 0\;
      }
      \Else{
       overtake\_status = 1\;
      }
        \KwRet overtake\_status;
    }
    
    \SetKwFunction{FMain}{Main}
    \SetKwProg{Fn}{Function}{:}{}
    \Fn{\FMain{}}{
    initialize: $\Pi = \{\}$ 
    \\
    
    \For{$\tau \in \mathbf{T}$}{
     initialize: p = \{\}, overtakes = \{\}, total = \{\} \\
     
     \For {$x,y,s \in \mathcal{X} X \mathcal{Y} X \mathcal{S}$}
     {
     initialize: $X_{ego}, X_{obs}$ \\
     \For{$t = 0$ to $ T_{sim}$}{
     $u^* = $  MPCC Planner ({$X_{obs}$}) \\
    Steer the ego: $X_{ego}^+ = f(X_{ego}, u^*)$ \\
    Update the obstacle position: $X_{obs}^+ = g(X_{obs})$ \\
    $X_{obs}, X_{ego} = X_{obs}^+, X_{ego}^+ $ \\
    Identify track portion $i \in \{0,1,2,3\}$ \\
    \uIf{Check Overtake($X_{obs}, X_{ego} $)}{
        $overtakes[\mathcal{R}_i]++$\; 
    }
      $total[\mathcal{R}_i] ++$ \;
     }
    
     }
     
     $p[\mathcal{R}_i] = overtakes[\mathcal{R}_i] /total[\mathcal{R}_i]$ \\
     Compute: $p[\mathcal{R}_1],p[\mathcal{R}_2],p[\mathcal{R}_3], p[\mathcal{R}_4]$
     $\Pi[\tau] = argmax(p)$}
     
     }
\end{algorithm}

%%%%%%%%%%%%%%%%%%%%%%%%%%%%%%%%%%%%%%%%%%%%%%%%%%%%%%%%%%%%%%%%%%%%%%%%%%%%%%%%
%%%%%%%%%%%%%%%%%%%%%%%%%%%%%%%%%%%%%%%%%%%%%%%%%%%%%%%%%%%%%%%%%%%%%%%%%%%%%%%%
%%%%%%%%%%%%%%%%%%%%%%%%%%%%%%%%%%%%%%%%%%%%%%%%%%%%%%%%%%%%%%%%%%%%%%%%%%%%%%%%

\subsection{ Offline Experiment Results}

The offline policy learning algorithm is run on the described experimental setup. Based on the outcome of the algorithms different results are shown. Firstly we display the results for one specific track portions. 

\begin{figure}[h!]
\begin{center}
    \includegraphics{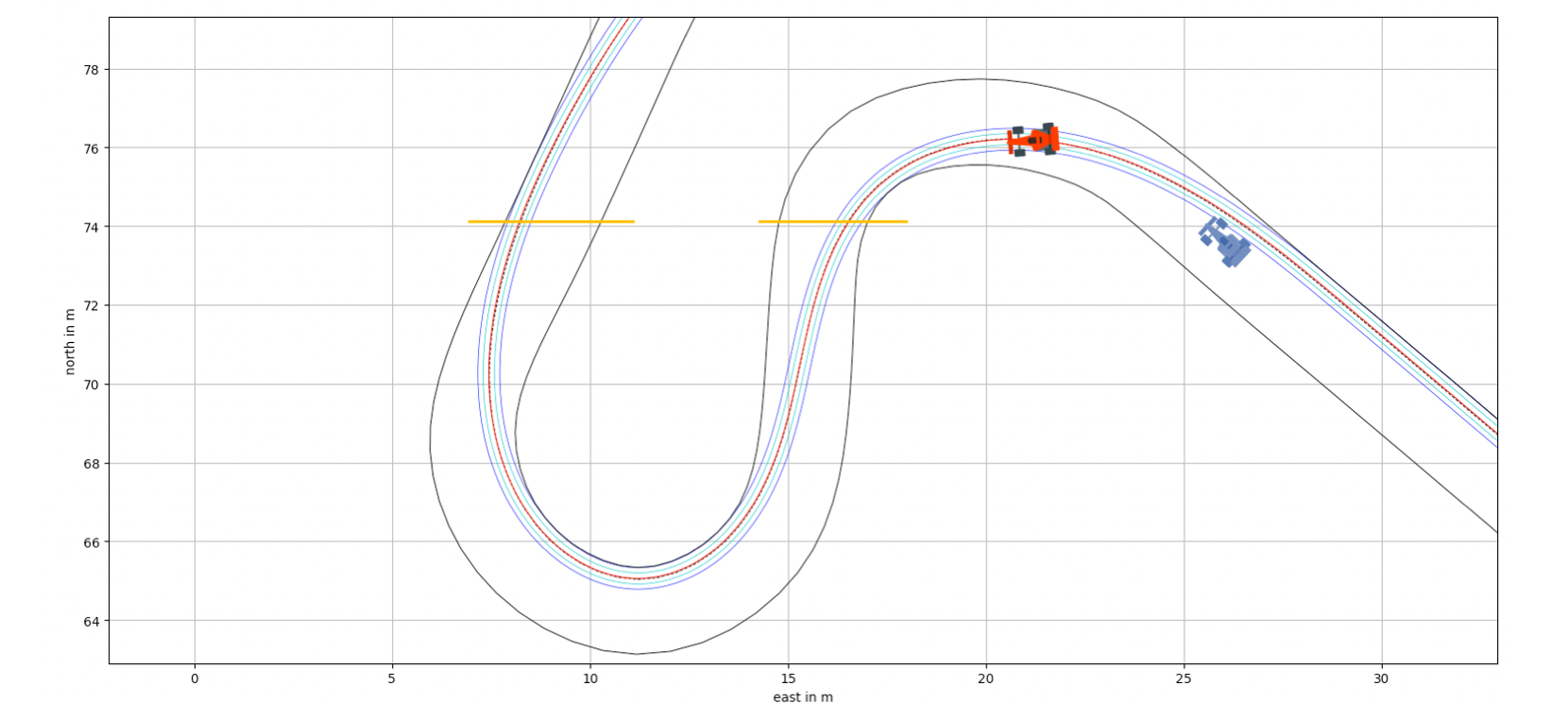}
    \caption{Hairpin Curve - Overtake analysis}
    \label{figure:hairpin}
    \end{center}
\end{figure}

We choose the hairpin (track portion label 3) which has a predefined overtaking window (marked in yellow). The obstacle vehicle (red) is tracking the optimal race line and the ego vehicle (blue) starts behind the obstacle (\ref{figure:hairpin})

\begin{figure}[h!]
\begin{center}
    \includegraphics[scale = 1.0]{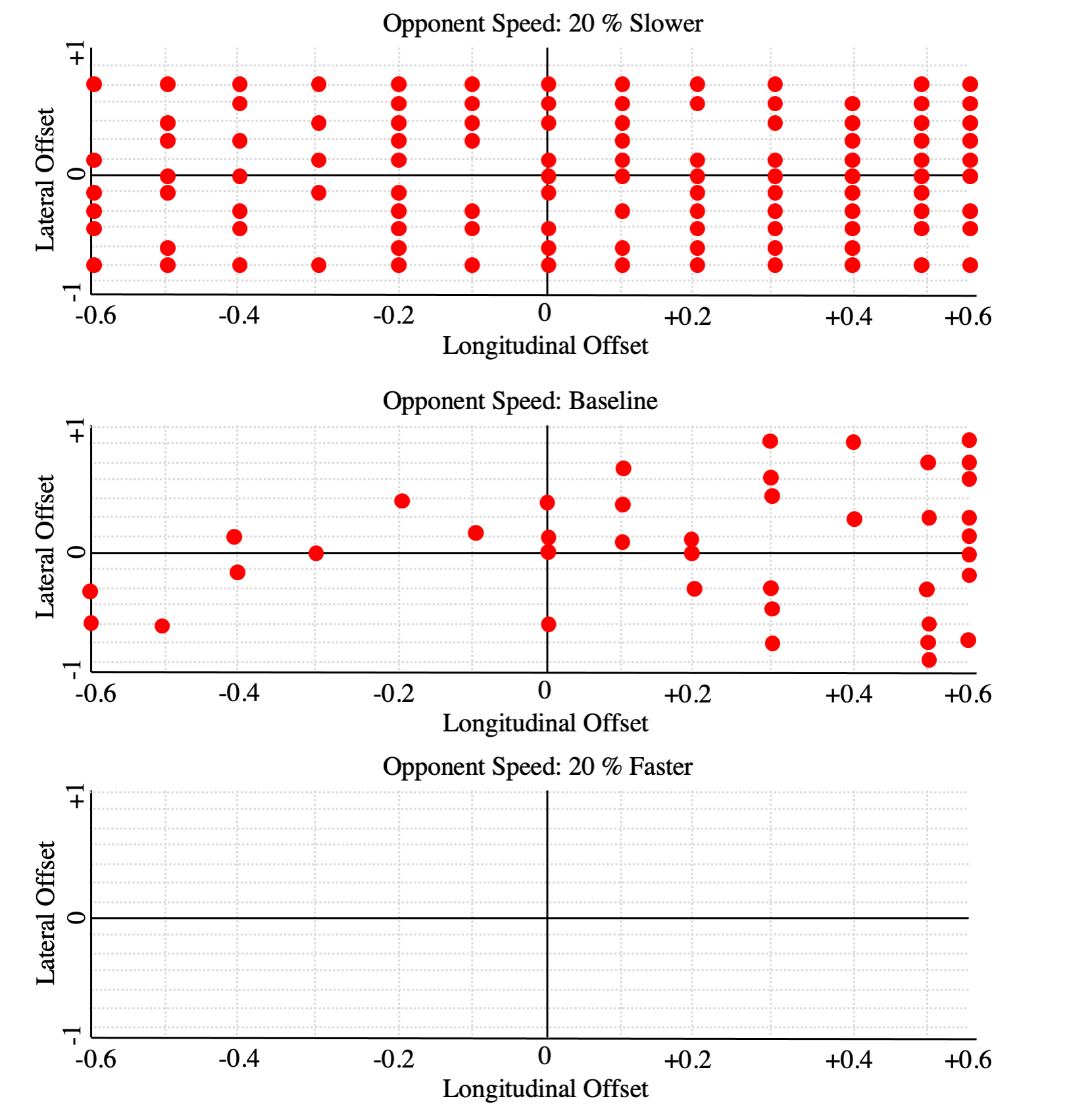}
    \caption{Display of successfull overtaking maneuvers over different positions across different speed offsets for the hairpin (track portion 3)}
    \label{figure:speed_vs_overtakes}
    \end{center}
\end{figure}

The ego vehicle is said to overtake the obstacle if it is ahead of the obstacle before the overtaking corridor ends. The overtaking corridor defines the portion of the track in which a successful overtake must happen. If the ego vehicle passes the obstacle vehicle outside of the corridor, it will not count as a successful overtake. Similar overtaking corridors are defined for all 8 track portions. As a results we get for each experiment the feedback if the overtaking maneuver was successful (figure \ref{figure:speed_vs_overtakes}).

The Algorithm \ref{alg:policyalg} is run for the track portion and the overtaking probability distributions are obtained. With this, the policy for this curve is also obtained. Table \ref{tab:hairpin_probab} below shows the probability distribution of overtakes at this curve. 
\begin{table}[h!]
 \caption{Overtaking Probabilities for Hairpin}
    \centering
    \begin{tabularx}{0.3\textwidth} { 
  | >{\centering\arraybackslash}X 
  | >{\centering\arraybackslash}X  | }
 \hline
 Region & Probability \\
 \hline
 $\mathcal{R}_1$  & 0.41  \\
\hline
 $\mathcal{R}_2$   & 0.38  \\
\hline
 $\mathcal{R}_3$   & 0.25  \\
\hline
 $\mathcal{R}_4$   & 0.21  \\
\hline
\end{tabularx}
\label{tab:hairpin_probab}
\end{table}

\begin{figure}[h!]
\begin{center}
    \includegraphics{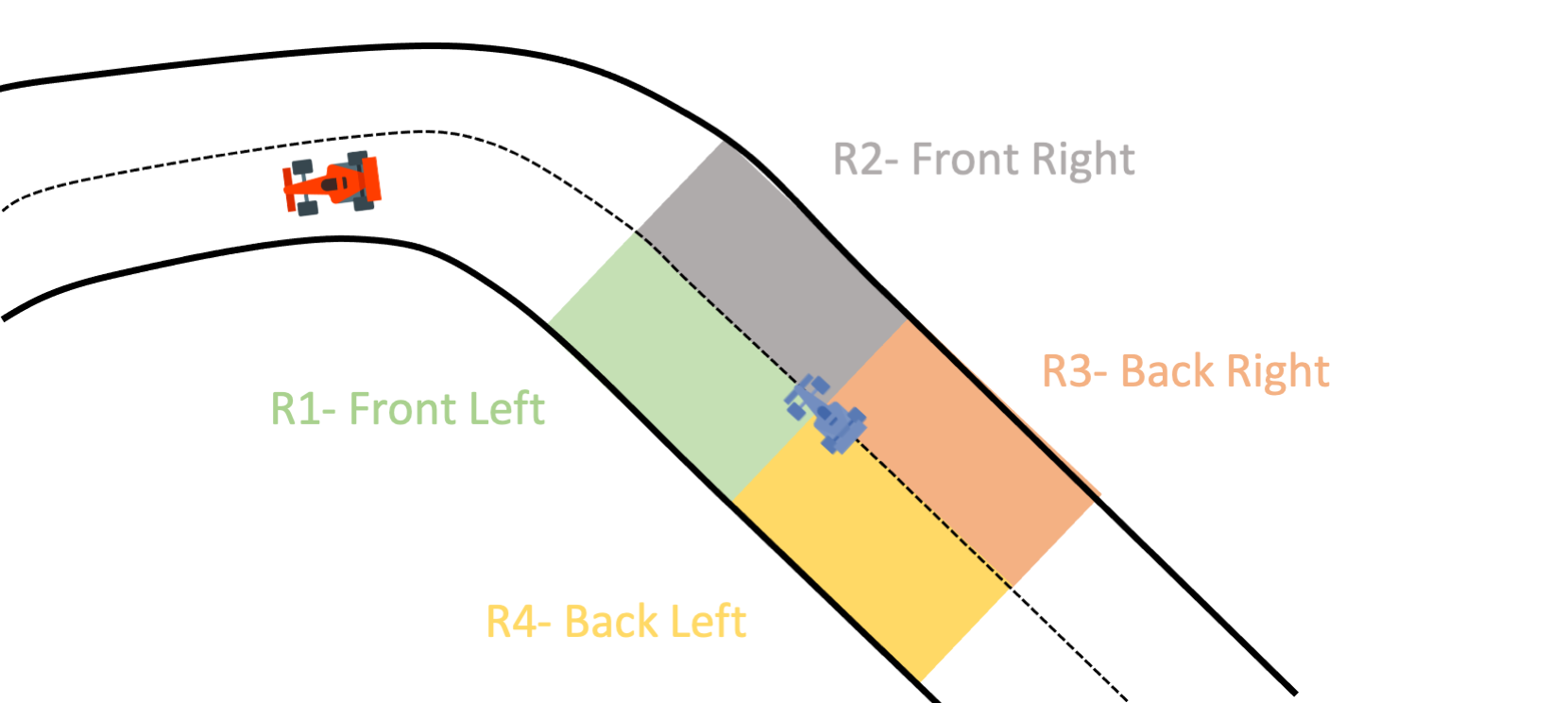}
    \caption{Hairpin curve split in four track regions of interest}
    \label{figure:track_regions}
    \end{center}
\end{figure}

To use this probability later for our switching MPCC algorithm we define these four regions in latent space on the track as displayed in figure \ref{figure:track_regions}.

The distribution of overtakes with respect to change in obstacle speed is shown in Figure \ref{figure:speed_vs_overtakes}. It can be observed that the number of overtakes is highest if the obstacle speed is 20\% lesser than its baseline. However, when there is no speed advantage (baseline scenario), we can observe that the region $\mathcal{R}_1$ has higher overtakes than any other region. Therefore, the policy for this track portion would ensure that the ego vehicle starts at $\mathcal{R}_1$ before entering the hairpin which gives it a positional advantage and increases it overtaking probability. This process is repeated for all track portions and the policies are learnt from the offline experiments. \newline

Besides the case study on the hairpin Table \ref{tab:all_probab} contains the overtaking probabilities for all the eight derived track portions. The policy map is developed using the results from the experiments.

\begin{table}[h!]
 \caption{Overtaking Probabilities for all track portions}
    \centering
    \begin{tabularx}{0.45\textwidth} { 
    | >{\centering\arraybackslash}X 
  | >{\centering\arraybackslash}X 
  | >{\centering\arraybackslash}X 
  | >{\centering\arraybackslash}X 
  | >{\centering\arraybackslash}X 
  | >{\centering\arraybackslash}X | }
 \hline
 Track Portion ($\tau$)& Type & $p(\mathcal{R}_1 )$ & $p(\mathcal{R}_2 )$ & $p(\mathcal{R}_3 )$ & $p(\mathcal{R}_4 )$ \\
 \hline
 1 & Sweeper & 0.99 & 0.93 & 0.52 & 0.59 \\
\hline
 2 & Hairpin & 0.63 & 0.52 & 0.31 & 0.33 \\
\hline
 3 & Hairpin & 0.41 & 0.38 & 0.25 & 0.21\\
\hline
 4 & Sweeper & 0.65 & 0.67 & 0.57  & 0.59\\
\hline
  5 & Chicane & 0.25 & 0.21 & 0.14 & 0.21\\
\hline
  6 & Straight & 0.99 & 1.0 & 0.95 & 0.99\\
\hline
  7 & Sweeper & 0.47 & 0.52 & 0.33 & 0.31\\
\hline
  8 & Hairpin & 0.40 & 0.36 & 0.37 & 0.38\\
\hline
\end{tabularx}
\label{tab:all_probab}
\end{table}

From the statistical data obtained from offline experiments, we have the following observations and outcomes:
\begin{itemize}

\item  On the straight, it does not really  matter where we are on the track when trying to overtake. Track region 3 (position: back right) is the one region with the lowest overtaking probability. This is due do the cause that the on the straight the optimal raceline leads from left to right across the straight. Starting in the back right would lead to an braking and left overtaking maneuver which is causing problems for turn 7 afterwards. \\
\item We can see that in each hairpin we have the highest overtaking probability in region 1 (position: front left). This is due to the fact that being on the inside in the hairpin that car is able to achieve a better trajectory inside the hairpin. \\
\item The sweeper curve generally has a high overtaking probability due to the high speeds of the car. We only get an advantage here if we are close enough to the car and therefore we need to be in region $p(\mathcal{R}_1)$ or $p(\mathcal{R}_2)$. In both track portions 4 and 7 we see a higher overtaking maneuver from starting in the front right track regions. \\
 \item The chicane has  generally a low overtaking probability due to the fact that it is a complex region to handle for the car and with little space for an overtaking maneuver. We see a highest overtaking probability here on the front left region and in contrary the lowest overtaking probability in the back right region. Both front right and back left region show the same overtaking probability and therefore being far behind the car lowers here again the overtaking possibility of the ego car. 

\end{itemize}

\subsection{Online switching MPCC Results}

Firstly we define the algorithm for the switching MPCC:

\begin{algorithm}
    \label{alg:racealg}
    \SetAlgoLined
    \SetKwFunction{FMain}{ Switched MPCC}
    \SetKwProg{Fn}{Function}{:}{}
    \Fn{\FMain{$X_{obs}$, mode}}{
    
     \uIf{mode = `normal`}{
        $u^* = $ Solve MPCC Problem in Section \ref{sec:optimalcontrol}
      }
      \Else{
       Modify MPCC to integrate policy (left,right) \\
        $u^* = $ Solve Modified MPCC Problem
       
      }

        \KwRet $u^*$\;
    }
    \textbf{initialize} $X_{ego}$, $X_{obs}$ mode = `normal` \\
     \For{$t = 0$ to $ T_{sim}$}{
     $u^* = $  Switched MPCC ({$X_{obs}$},mode) \\
    Steer the ego: $X_{ego}^+ = f(X_{ego}, u^*)$ \\
    Update the obstacle position: $X_{obs}^+ = g(X_{obs})$ \\
    $X_{obs}, X_{ego} = X_{obs}^+, X_{ego}^+ $ \\
    Identify track portion  $\tau$ where ego is present \\
    Policy lookup: mode = $\Pi[\tau]$
    }
\end{algorithm}

With the results of the track region probability we now have the possibility to use the policies obtained from algorithm \ref{alg:policyalg}  in our switching MPCC algorithm. We use this information to make the car move to the regions of high overtaking probability as displayed before

We repeat the experiments at all track portions with the learnt policies and compute the number of successful overtakes. Table \ref{tab:num_overtakes} summarizes the number of overtakes before and after the policy is integrated into the ego vehicle's MPCC planner. For every track portion, there are 576 trajectories owing to $\mathcal{X} = \{-0.8,-0.7,-0.6,-0.5,-0.4,-0.3,-0.2,-0.1,0,0.1,\\
0.2,0.3,0.4,0.5,0.6,0.7,0.8\}$, $\mathcal{Y} = \{-0.6,-0.5,-0.4,\\ -0.3,
-0.2,-0.1,0,0.1,
0.2,0.3,0.4,0.5,0.6\}$ and $\mathcal{S} = \{-0.2,0,0.2\}$. \newline

\begin{table}[h!]
 \caption{Number of overtakes with and without policy}
    \centering
    \begin{tabularx}{0.4\textwidth} { | >{\centering\arraybackslash}X | >{\centering\arraybackslash}X  | >{\centering\arraybackslash}X | >{\centering\arraybackslash}X  | }
 \hline
 Track Portion ($\tau$) & Track Portion Type & Number of Overtakes  Policy OFF & Number of Overtakes  Policy ON \\
 \hline
 1 & Sweeper & 436 & 452 \\
\hline
 2 & Hairpin & 256 & 337 \\
\hline
 3 & Hairpin & 308 & 426\\
\hline
 4 & Sweeper & 342 & 357 \\
\hline
  5 & Chicane  & 117 & 302 \\
\hline
  6 & Straight & \emph{565} & \emph{566}\\
\hline
  7 & Sweeper & 237 & 283\\
\hline
  8 & Hairpin & \textbf{218} & \textbf{394}\\
\hline
\end{tabularx}
\label{tab:num_overtakes}
\end{table} 

The results display that the offline policy learning approach is successful and leads to an increased number of overtaking maneuvers when implemented in a path planning approach. For all eight track portions we see more overtaking maneuvers. \\
We can observe that overtaking on the straights usually easy (even without policy) and therefore we don not see many more overtaking maneuvers with switching policy. The same is for the sweeper curve. Since sweeper curves are usually wide track portions that allow high speeds and not complicated steering maneuver, both with and without switching policy achieve high maneuvers. Although we see that the switching policy leads to more overtaking maneuvers. \\
We see the biggest impact in both the hairpins and the chicane. This is mainly due to the fact, that overtaking at these hairpins is usually complicated and needs a good strategy beforehand. We see that we can nearly triple the amount of overtaking maneuvers in the chicane which shows, that having the right starting position for an overtaking maneuver is indispensable. The best results in this experiment is achieved in track portion 8 (hairpin) where a total of 186 more overtakings are done with the MPCC switching.\\
In summary we see that the gains for the overtaking maneuver are less for straight sections such as straights and sweeper curve and better for curved sections such as the hairpin and the chicane.

%%%%%%%%%%%%%%%%%%%%%%%%%%%%%%%%%%%%%%%%%%%%%%%%%%%%%%%%%%%%%%%%%%%%%%%%%%%%%%%%
%%%%%%%%%%%%%%%%%%%%%%%%%%%%%%%%%%%%%%%%%%%%%%%%%%%%%%%%%%%%%%%%%%%%%%%%%%%%%%%%
%%%%%%%%%%%%%%%%%%%%%%%%%%%%%%%%%%%%%%%%%%%%%%%%%%%%%%%%%%%%%%%%%%%%%%%%%%%%%%%%

\section{Conclusion and Future Work}
\label{sec:discussion}

In this paper, a track based policy learning from offline experiments is proposed to learn effective overtaking strategies based on position advantage at different track portions. A switched Model Predictive Contouring Control (MPCC) scheme was proposed to integrate driving behaviours/policies into the motion planning and control of the vehicle. Extensive simulations on real world racetrack layout with a naive, non-interactive obstacle shows that the offline policy learning algorithm is able to provide areas for high overtaking maneuvers. \\
By integrating these areas in a switching MPCC method we could show that the policy based Switching MPCC approach has more overtakes than the regular MPCC planner. Some track portions like straight have not shown considerable increase in the number of overtakes, as the overtaking probabilities are already highest (around 0.99) in a straight-way. However, the policy based algorithm was found to be highly effective for convoluted track portions like chicanes, where a positional advantage plays a major role in successful overtaking maneuvers. \\
The offline policy learning shows promising results for a naive opponent following a pre-computed raceline with no interactions. Future directions of research include extending the policy learning algorithm to race against a variety of opponent types. We consider: non-reactive, reactive and aggressive opponents which are defensive and sophisticated to overtake. Other dimensions of research include the evaluating the policy based algorithm across different tracks and study the effect of changing track parameters on track-policy based overtaking.

%%%%%%%%%%%%%%%%%%%%%%%%%%%%%%%%%%%%%%%%%%%%%%%%%%%%%%%%%%%%%%%%%%%%%%%%%%%%%%%%
%%%%%%%%%%%%%%%%%%%%%%%%%%%%%%%%%%%%%%%%%%%%%%%%%%%%%%%%%%%%%%%%%%%%%%%%%%%%%%%%
%%%%%%%%%%%%%%%%%%%%%%%%%%%%%%%%%%%%%%%%%%%%%%%%%%%%%%%%%%%%%%%%%%%%%%%%%%%%%%%%

\printbibliography

\section{Appendix}
\subsection{Vehicle Model}
\label{sec:vehiclemodel}

The simulation is built around the dynamic model used in \cite{liniger2015optimization} and is implemented with
adapted parameters. It is a single track bicycle model with pajecka tire forces.
\begin{figure}[h!]
\includegraphics[scale = 0.37]{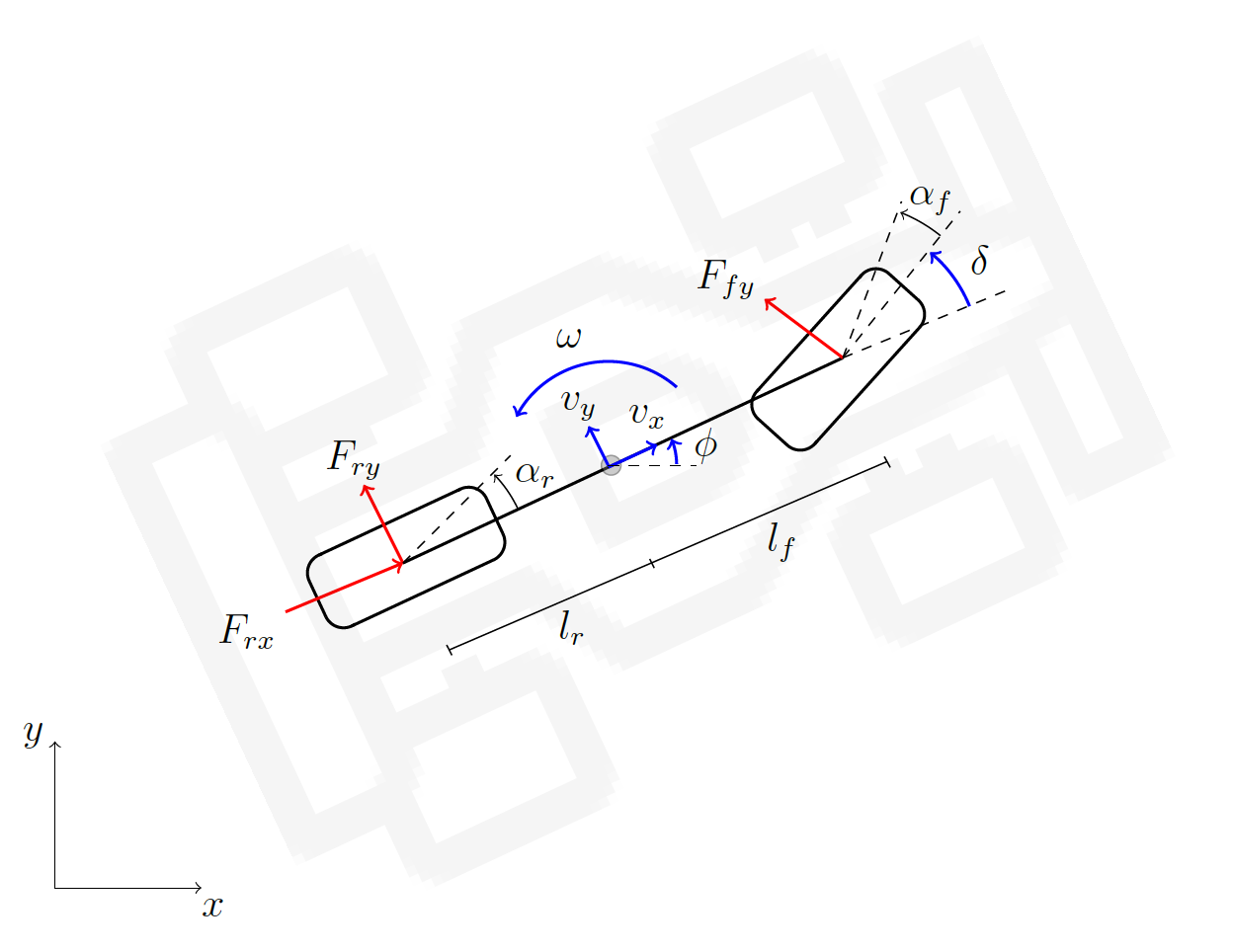}
\caption{Single track model with pajecka tire forces}
\label{fig:bicycle_model}
\end{figure}
The single-track bicycle model is used for modelling the race car (ego-vehicle). The model is based on in-plane motions and pitch, roll as well as load-changes are neglected. Due to the rare-wheel drive nature and absence of active breaking, longitudinal forces on the front wheel are also neglected. 

The dynamics of this model read:
$$
\dot{X} =v_{x} \cos (\phi)-v y \sin (\phi) 
$$

$$
\dot{Y} =v_{x} \sin (\phi)+v y \cos (\phi) 
$$
$$
\dot{\phi} =\omega 
$$
$$
\dot{v}_{x} =\frac{1}{m}\left(F_{r x}-F_{f y} \sin (\delta)+m v_{y} \omega\right) 
$$
$$
\dot{v}_{x} =\frac{1}{m}\left(F_{r y}+F_{f y} \cos (\delta)-m v_{x} \omega\right) $$
$$
\dot{\omega} =\frac{1}{I_{z}}\left(F_{f y} l_{f} \cos (\delta)-F_{r y} l_{r}\right) $$
$$
\dot{d} =u_{\dot{d}} $$
$$
\dot{\delta} =u_{\dot{\delta}} $$
$$
\dot{\theta} =u_{\dot{\theta}}
$$
where
$$
\alpha_{f} =-\arctan \left(\frac{\omega l_{f}+v_{y}}{v_{x}}\right)+\delta $$
$$
\alpha_{r} =\arctan \left(\frac{\omega l_{r}-v_{y}}{v_{x}}\right) $$
$$
F_{f y} =D_{f} \sin \left(C_{f} \arctan \left(B_{f} \alpha_{f}\right)\right) $$
$$
F_{r y} =D_{r} \sin \left(C_{r} \arctan \left(B_{r} \alpha_{r}\right)\right) $$
$$
F_{r x} =\left(C_{m 1}-C_{m 2} v_{x}\right) d-C_{r o l l}-C_{d} v_{x}^{2}
$$ 

$X \in \mathcal{X}$ : 
$X = [x, y, \phi, v_x, v_y, \omega, d, \delta, \theta]$ is the state of the car. 
$x,y$ are position of the centre of gravity, $\phi$ is the yaw angle of the vehicle with reference to world frame, $v_x, v_y$ are longitudinal and lateral velocities and $\omega$ is the yaw rate. \\
The parameter $\theta$ is known as the advancing parameter (which is augmented to form the complete state vector), $d$ is the integrated motor torque and $\delta$ is the steering angle. \\

$u \in \mathcal{U}$: $u =[u_{\dot{d}},u_{\dot{\delta}},u_{\dot{\theta}}]$ is the input to the model. These inputs are the derivatives of the commandable inputs to the vehicle chosen so as to penalise their smoothness. \\

$\mathcal{X} \subset \mathbf{R}^9$ and $\mathcal{U} \subset \mathbf{R}^3$ are the set of admissible states and control inputs for the vehicle defined by the optimal control problem in Section \ref{sec:optimalcontrol}

\end{document}